\documentclass[letterpaper, 10 pt, conference]{ieeeconf}  
\IEEEoverridecommandlockouts                              

\usepackage{graphicx}
\usepackage{multirow}
\usepackage{comment}
\usepackage{amsmath}
\usepackage{cleveref}
\usepackage{amssymb}
\usepackage{cite}
\usepackage{array}
\usepackage{booktabs}
\newcommand{\ra}[1]{\renewcommand{\arraystretch}{#1}}

\title{\LARGE \bf
   Developments in Modern GNSS 
   \\
   and Its Impact on 
   \\
Autonomous Vehicle Architectures
}

\author{Niels Joubert, Tyler G. R. Reid, and Fergus Noble
\thanks{Niels Joubert and Fergus Noble are with Swift Navigation, San Francisco, CA 94103, email: 
    {\tt\small njoubert@gmail.com, fergus@swift-nav.com}}%
\thanks{Tyler G.R. Reid is with Xona Space Systems, Vancouver, BC, email: 
    {\tt\small tyreid@alumni.stanford.edu}}%
}

\begin{document}

\maketitle
\thispagestyle{empty}
\pagestyle{empty}

\begin{abstract}
This paper surveys a number of recent developments in modern Global Navigation Satellite Systems (GNSS) and investigates the possible impact on autonomous driving architectures. Modern GNSS now consist of four independent global satellite constellations delivering modernized signals at multiple civil frequencies. New ground monitoring infrastructure, mathematical models, and internet services correct for errors in the GNSS signals at continent scale. Mass-market automotive-grade receiver chipsets are available at low Cost, Size, Weight, and Power (CSWaP). The result is that GNSS in 2020 delivers better than lane-level accurate localization with 99.99999\% integrity guarantees at over 95\% availability. In autonomous driving, SAE Level 2 partially autonomous vehicles are now available to consumers, capable of autonomously following lanes and performing basic maneuvers under human supervision. Furthermore, the first pilot programs of SAE Level 4 driverless vehicles are being demonstrated on public roads. 

However, autonomous driving is not a solved problem. GNSS can help. Specifically, incorporating high-integrity GNSS lane determination into vision-based architectures can unlock lane-level maneuvers and provide oversight to guarantee safety. Incorporating precision GNSS into LiDAR-based systems can unlock robustness and additional fallbacks for safety and utility. Lastly, GNSS provides interoperability through consistent timing and reference frames for future V2X scenarios.
\end{abstract}


\section{Introduction}

\begin{figure}[h]
    \centering
    \includegraphics[width=3.25in]{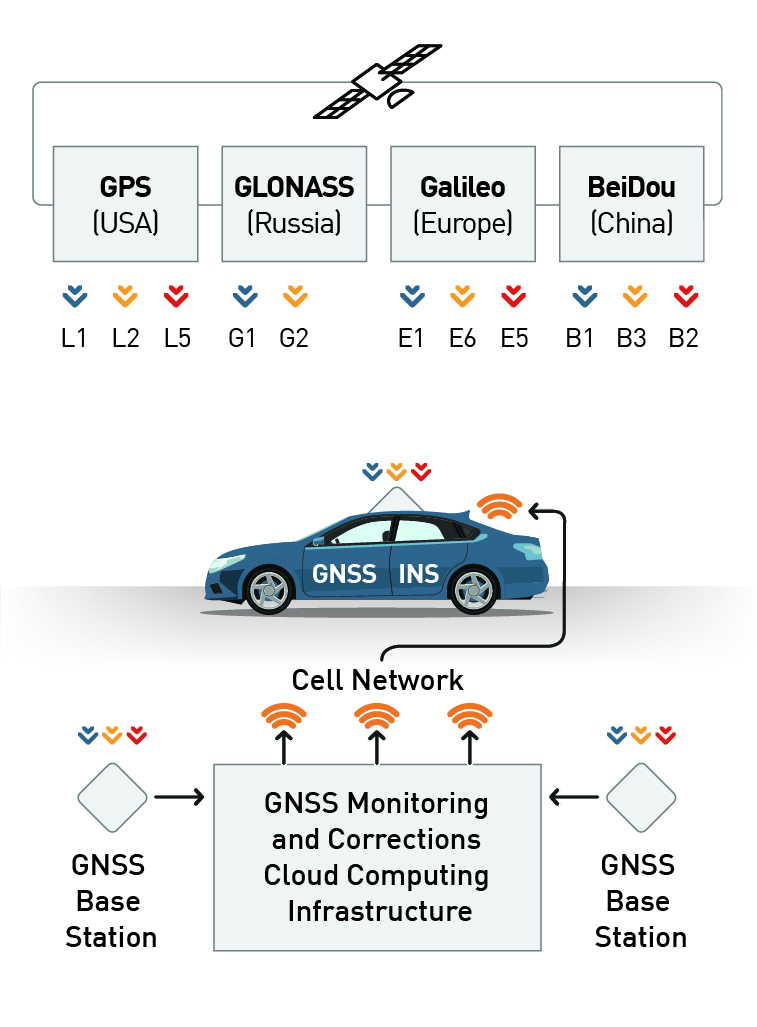}
  \caption{The modern GNSS automotive ecosystem. Vehicles equipped with automotive-grade low CSWaP hardware receives signals from four satellite constellations across three bands. Sparse ground station network backed by cloud computing provide error corrections and fault monitoring, delivered using standardized protocols via cellular networks.}
  
  \label{fig:gnss-end-to-end}
\end{figure}

Localization is a foundational capability of autonomous driving architectures. Knowledge of precise vehicle location, coupled with highly detailed maps (often called High Definition (HD) maps), add the context needed to drive with confidence. To maintain the vehicle within its lane, highway operation requires knowledge of location at 0.50 meters whereas local city roads require 0.30 meters~\cite{Reid2019c}. The challenge facing auto makers is meeting reliability at an allowable failure rate of once in a billion miles for Automotive Safety Integrity Level (ASIL) D~\cite{Reid2019c, Kafka2012}. Achieving this for autonomous vehicles has not yet been demonstrated. 

There are six levels (0 to 5) of autonomous driving as defined by the Society of Automotive Engineers (SAE)~\cite{SAEInternational2018}. Here, we explore current autonomous vehicle architectures for driver-supervised partial autonomy (SAE Level 2), and driverless operation in a restricted operating domain (SAE Level 4). Both approaches rely on perception sensors to understand the environment in which the vehicle must make driving decisions. Unfortunately, purely perception-based approaches struggle to fully solve the driving problem due to outages from environmental effects, faults from sensor glitches, and ambiguities in the real world. For instance, Google (Waymo) famously demonstrated the challenge of correctly interpreting an upside-down stop sign sticking out of the backpack of a cyclist~\cite{Anguelov2019}. The industry has looked toward robust localization systems and detailed maps to address these challenges.

GNSS and automated driving have a long lineage. Both have seen revolution. Early GNSS suffered from low accuracy, limited availability, and a lack of integrity. Still, the ability to globally localize on a map was already valuable enough that contenders in the 2005-2007 DARPA Challenges~\cite{Khatib2008} used GNSS, but only for road-level routing. Since then, GNSS experienced a step-change in performance and capabilities thanks to the development of the ecosystem shown in Figure~\ref{fig:gnss-end-to-end}. Simultaneously, autonomous driving has moved from a science experiment to driver-supervised consumer products and early driverless pilot programs. Yet reaching the safety, comfort, cost and utility levels required for widespread adoption of autonomous driving have proved slow and elusive. In this paper, we explore the potential role of a modern GNSS in achieving the ultimate autonomous driving safety goal of only one localization failure per billion miles per vehicle.

\section{Recent Developments in Modern GNSS} 
\label{sec:recent-gnss-dev}

In the last 20 years, key areas of development in modern GNSS have progressed performance to the point where decimeter-level accuracy with over 95\% availability is available for automotive. Furthermore, these developments have unlocked integrity capabilities --- the ability of the receiver to provide a trustworthy alert when it cannot guarantee the validity of its outputs to an extremely high probability~\cite{Elkaim2015}. We now survey seven key areas of development.

\subsection{Multiple Independent GNSS Constellations}

\begin{figure}[h]
    \centering
    \includegraphics[width=3.5in]{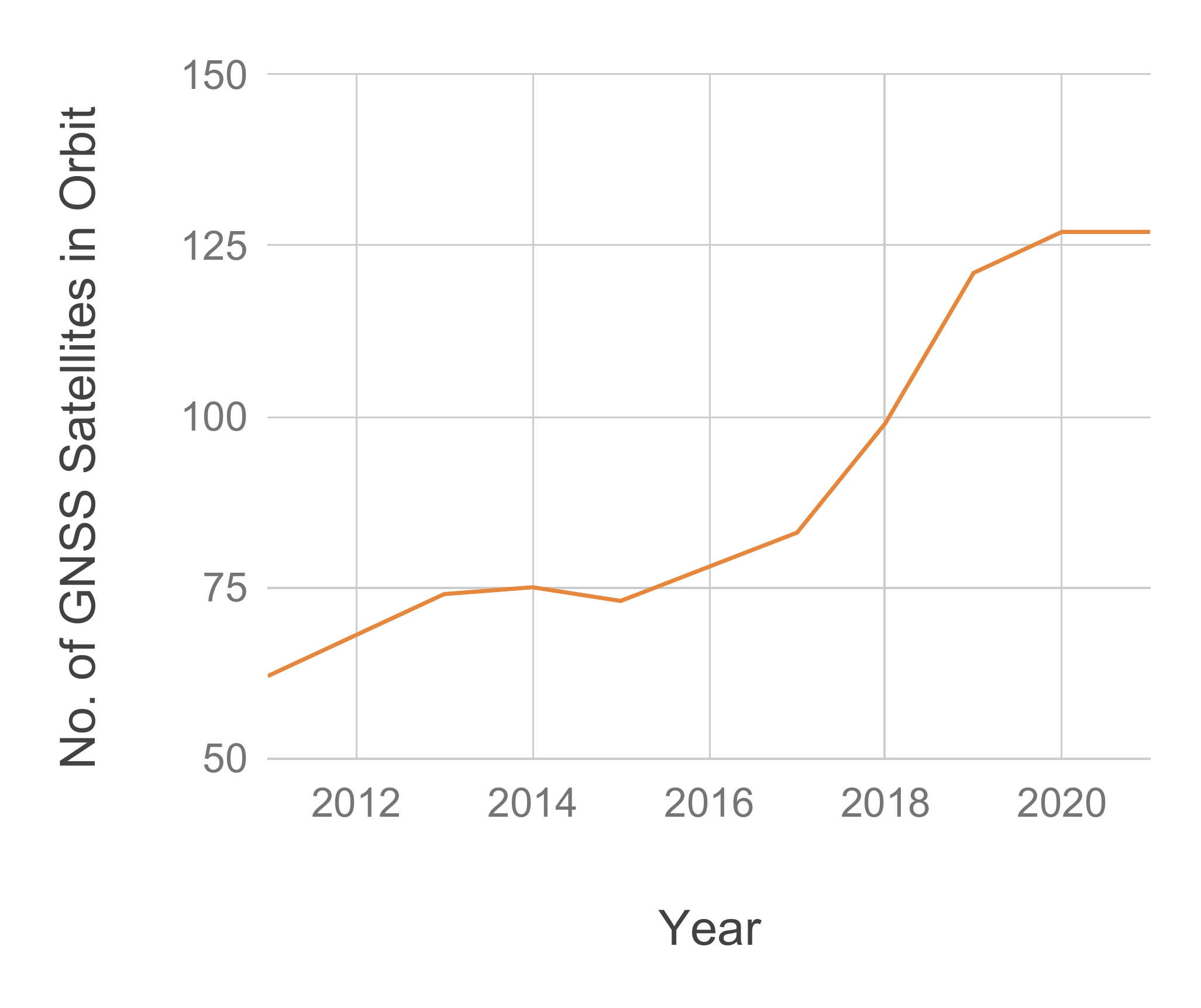}
  \caption{Number of GNSS satellites in orbit as a function of time. In 2005, GPS-only receivers had access to less than 30 satellites, while modern GNSS receivers have access to over 125.}
  \label{fig:gnss-new-sats}
\end{figure}

The U.S. GPS was declared fully operational in 1995 with 24 satellites providing global coverage, and opened for civil use in 2000. At that time, accuracy was around 10~meters~\cite{Ochieng2002}. Since then, three new global satellite navigation systems have been put into service by other nation-states, with more than 125 navigation satellites operational as of 2020. The progression is shown in Figure \ref{fig:gnss-new-sats}, highlighting that most of this increase happened since 2016. 

Russia's GLONASS was the second GNSS constellation to reach operational status, gaining market adoption and continuous global service in 2011~\cite{Langley2017}. China's BeiDou-3 operationalized 19 satellites starting in 2012, with the full 24 satellite constellation targeted for operational readiness by the end 2020~\cite{Shen2019}. The European Union's Galileo system operationalized 22 satellites starting in 2013 with an intended 24 satellites plus 6 spares expected to be completed by the end of 2020~\cite{Chatre2019}. There are already more than 1 billion Galileo-enabled devices in service. In addition to the four global systems, there are multiple regional systems coming online including Japan's 4-satellite Quazi-Zenith Satellite System (QZSS) and India's 7-satellite Navigation with Indian Constellation (NAVIC). QZSS is particularly interesting, since it is designed to provide coverage in dense urban areas, and transmits correction information to improve accuracy and provide basic integrity monitoring for GNSS~\cite{Sakai2019}. 

Most GNSS techniques work with as few as 5 satellites. Soon most users will have over 25 satellites above the horizon at all times. This redundancy is important for a number of reasons. The large number of satellites increases GNSS availability, since local obstructions can now block significant parts of the sky without preventing GNSS functionality. Indeed, Heng et al. demonstrated that a three-constellation system that can only see satellites more than 32 degrees above the horizon is equivalent in performance to a GPS-only constellation in an open-sky environment~\cite{Heng2014}. Furthermore, these satellite navigation systems are independently developed and operated, enabling integrity guarantees through cross-checking between constellations.

\subsection{Modern Signals Across Multiple Frequencies}

New GNSS satellites broadcast modernized signals on multiple civil frequencies. GPS, GLONASS, Galileo, BeiDou, and QZSS all operate in the legacy L1 / E1 / B1C band (around 1575.42~MHz). Further, all also operate or have plans to operate in the modernized L5 / E5a / B2a band (around 1176.45~MHz)~\cite{GPSDirectorate2019, EuropeanUnion2015, ChinaSatelliteNavigationOffice2017, CabinetOfficeGovernmentofJapan2018,Lu2019}.  GPS and GLONASS also operate in the L2 band (around 1227.6~MHz).

\begin{figure}[h]
    \centering
    \includegraphics[width=3.25in]{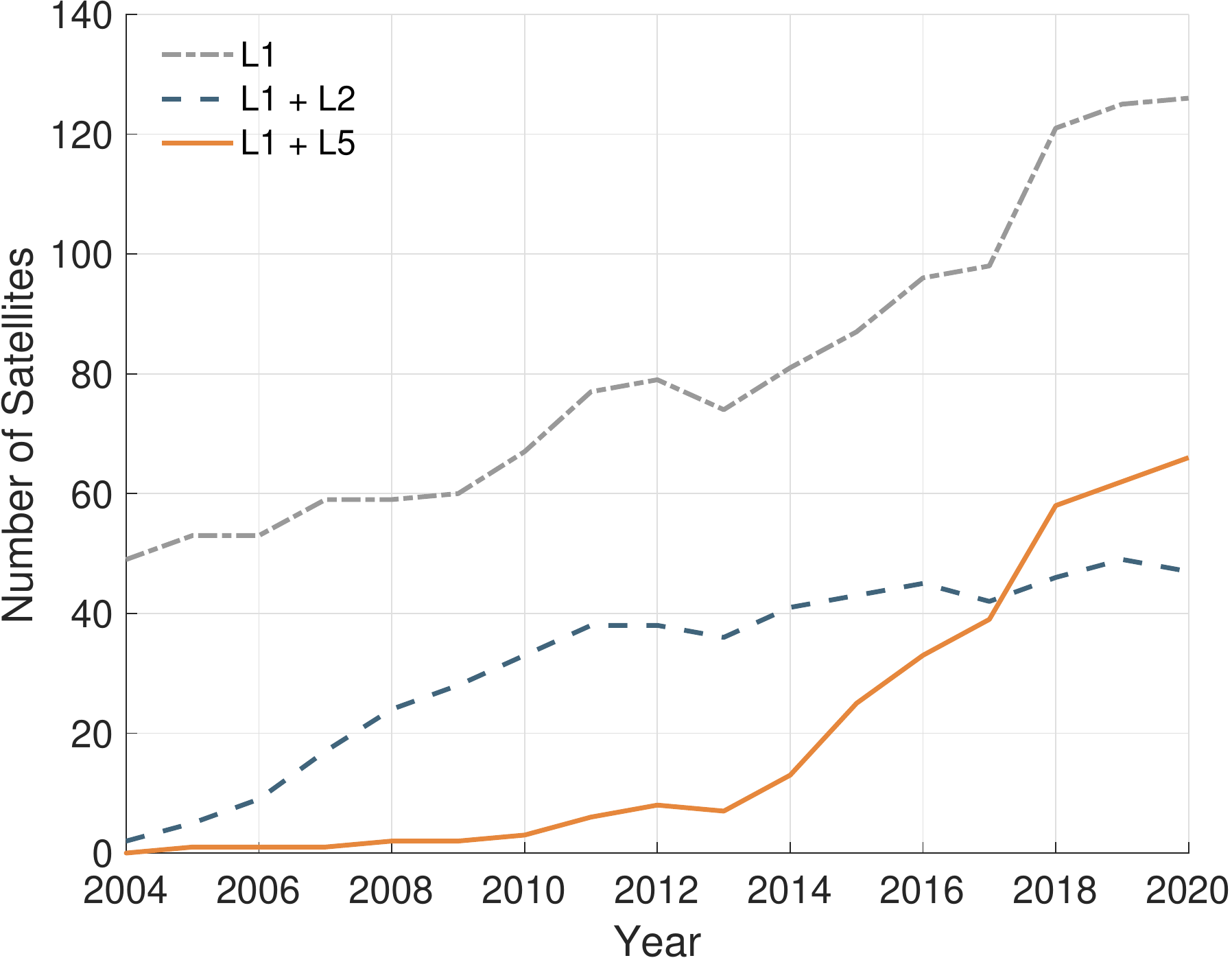}
  \caption{L1 (E1, B1C), L2, and L5 (E5a, B2a) civil signals as function of time. Today, more than half of GNSS satellites transmit on multiple bands.}
  \label{fig:gnss-new-sigs}
\end{figure}

The L5 band offers tenfold more bandwidth than the L1 signal. Modern GNSS signals use the additional bandwidth to offer up to an order of magnitude better satellite ranging precision, improved performance in multi-path environments, protection against narrowband interference, and speeding up signal acquisition from 2~seconds to 200~milliseconds~\cite{Hegarty2008, Tran2004}. Furthermore, the L5 band is reserved for safety-of-life applications, enabling strict regulations against radio frequency interference. This is helpful, since several devices in the autonomous vehicle world have been known to cause interference including USB 3.0 connections with L2. Figure \ref{fig:gnss-new-sigs} shows the progress in the number of multi-frequency satellites available. 

\subsection{Error Correction Algorithms for High Precision GNSS}

The accuracy of standard GNSS is degraded due to noise and biases in the satellite orbits and clocks, hardware, atmospheric conditions, and other effects. Similarly, standard GNSS has no provisions to protect against faults. High precision GNSS deploy error correction and fault detection algorithms to provide up to centimeter-level accuracy and integrity guarantees. Table \ref{tab:corrections-sum} shows the three major techniques employed in achieving GNSS precision: RTK, PPP, and the hybrid PPP-RTK.

The Real-Time Kinematic (RTK) method calculates a centimeter accurate baseline to a local static reference receiver by differencing the received signals between both receivers. Differencing cancels common mode signals, producing an accurate 3D vector between the static and roving receiver as long as the two receivers are within a few dozen kilometers. This approach resolves what is known as the carrier phase integer ambiguity. Performing integer ambiguity resolution allows using only the carrier phase of the GNSS signal for positioning, which provides centimeter accurate satellite ranging. RTK provides the highest accuracy of precision methods, but requires a dense monitoring network to cover a large area, complex handoff procedures between base stations~\cite{Murrian2016a}, and does not natively provide integrity guarantees.

\begin{table}[h]
    \centering
      \caption{
Error Correction approaches for GNSS.
  }
	\begin{tabular}{c c c c}

		\toprule
         &
        PPP &
        RTK &
        PPP-RTK \\
		\midrule
        
        Accuracy &
        0.30~m &
        0.02~m &
        0.10~m \\
        
        Convergence Time &
        \textgreater 10~minutes &
        20~seconds &
        20~seconds \\
        
        Coverage &
        Global &
        Regional &
        Continental \\
        
        Seamless &
        Yes &
        No &
        Yes \\
        
		\bottomrule
	\end{tabular}
  \label{tab:corrections-sum}
\end{table}

The Precise Point Positioning (PPP)~\cite{Kouba2001} technique corrects individual errors by utilizing precise orbit and clock corrections, estimated from less than 100 global reference receivers. This technique scales globally, but requires the receiver to estimate the atmospheric errors slowly over time, taking many minutes to converge. PPP approaches do not use carrier-only ranging, and are typically limited to decimeter accuracy for automotive.

Modern GNSS has developed the hybrid PPP-RTK method to overcome the limitations of PPP and RTK~\cite{Wubbena2005}. Similar to PPP, it uses a network of ground stations to estimate errors directly. Similar to RTK, it solves the integer ambiguity problem to find centimeter-accurate ranges to satellites. This approach has recently been shown to be viable with \textgreater 150~km spacing between GNSS base stations~\cite{Banville2018}. PPP-RTK is attractive since this approach can scale corrections to continent-level seamlessly without the challenges of an RTK approach, and can achieve meter-level protection levels with formal guarantees on integrity to the level of 10\textsuperscript{-7} probability of failure per hour or a reliability of 99.99999\%~\cite{Gunning2018}.

\subsection{Ground-based GNSS Monitoring Networks}

Calculating corrections and performing fault detection requires ground monitoring networks. Several players are now deploying large-scale correction services targeted at automotive applications in North America and Europe including Hexagon~\cite{Jokinen2018}, Sapcorda~\cite{Vana2018}, Swift Navigation~\cite{GPSWorldStaff2019}, Trimble~\cite{Weinbach2018}, in China with players like Qianxun~\cite{Chen2018a}, and in Japan with the state-sponsored QZSS service for automated driving~\cite{Asari2016}. These networks often deploy PPP-RTK style approaches that additionally include specialized integrity monitoring solutions~\cite{GPSWorldStaff2019,EuropeanGlobalNavigationSatelliteSystemsAgency2019}.

\subsection{GNSS Corrections Data Standardization}

Corrections data have to be delivered to receivers in an understandable format. Demand from the automotive and cellular industries is leading to interoperability between networks and devices~\cite{Vana2018, Vana2019}, driving standardization of corrections data. Most relevant to the automotive community, The 3rd Generation Partnership Project (3GPP) is integrating GNSS corrections data directly into the control plane of the 5G cellular data network~\cite{3GPPOrganizationalPartners2019}. This integration allows broadcasting standardized corrections data to all vehicles simultaneous for lower cost, higher reliable, and better scalability than point-to-point connections. A variant of this approach is also available as the Centimeter Level Augmentation Service (CLAS) broadcast from the QZSS satellites over Japan. Interoperability and standardization enables the globalization of high-accuracy high-integrity positioning networks as a service.

\subsection{New Geodetic Datums \& Earth Crustal Models}

A challenge facing precision applications is the constant movement of the Earth's crust. In California's coast, tectonic shift is as much as 0.10~m a year laterally~\cite{Zeng2016}. Tidal forces due to the Moon and Sun deform the Earth's surface by as much as 0.40~m over six hours~\cite{Baker1984}. The weight of ocean tides can result in a further 0.10~m of deformation~\cite{Baker1984}. High accuracy reference frames for maps and GNSS must account for these effects. Fortunately, the modern ITRF2014 and ITRF2020~\cite{Altamimi2018} datums and services such as NOAA's Horizontal Time-Dependent Positioning~\cite{Pearson2013} can do so, keeping maps and GNSS localization consistent for decades even across continents.

\subsection{Mass-Market Automotive GNSS Chipsets}

The improvements in GNSS satellites is only beneficial if capable and affordable receiver hardware exists. Fortunately, multi-frequency, multi-constellation mass-market ASIL-certified GNSS chipsets are now available. Automotive-grade dual-frequency receivers were first showcased in 2018~\cite{TracyCozzens} delivering decimeter positioning~\cite{DeGroot2018a}. Major players in this space now include STMicroelectronics with its Teseo APP and Teseo V~\cite{TracyCozzens}, u-blox with its F9~\cite{Cozzens2018}, and Qualcomm with its Snapdragon~\cite{Qualcomm2019}. The CSWaP of these units are mostly \textless \$10 and \textless 10~Watt with a total system footprint of \textless 0.1~m\textsuperscript{2}. Moreover, many of these devices are ASIL-capable, enabling a positioning solution compliant with ISO 26262 automotive safety standards, meeting and exceeding current autonomy requirements~\cite{TracyCozzens, GPSWorldStaff2019}.

\section{On-Road GNSS Performance}

\begin{table*}[h]
  \centering
  \caption{Select data points that show on-road GNSS performance improvements between 2000-2019.} 
  \label{tab:gnss-perf-sum}
	\ra{1.5}
	\begin{tabular}
	{>{\centering\arraybackslash}m{0.5in}
	>{\centering\arraybackslash}m{0.3in} 
	>{\centering\arraybackslash}m{0.6in}
	>{\centering\arraybackslash}m{0.3in}
	>{\centering\arraybackslash}m{0.4in}
	>{\centering\arraybackslash}m{0.4in}
	>{\centering\arraybackslash}m{0.5in}
	>{\centering\arraybackslash}m{0.5in}
	>{\centering\arraybackslash}m{0.6in}
	>{\centering\arraybackslash}m{0.6in}
	>{\centering\arraybackslash}m{0.6in}}

		\toprule
		Source & 
		Year of Data & 
		Data Set &	
		Const. & 
		Freq. & 
		Receiver Type & 
		GNSS Corrections & 
		Env. & 
		Accuracy & 
		Availability & 
		Outage Times \\
		
		\midrule

		\cite{Ochieng2002} & 
		2000 & 
		2 hours & 
		GPS & 
		L1 & 
		Survey & 
		None & 
		Urban &
		10m, 74\%, Lateral & 
		28\% & 
		4.7~min, Worst-Case 
		\\
		
		\cite{Pilutti2010} & 
		2010 & 
		186 hours (13,000 km) & 
		GPS & 
		L1 & 
		Survey & 
		None & 
		Urban, Suburban, Rural, Highway &
		- & 
		85\%, Code Phase Position (HDOP $>$ 3) & 
		28 sec, 95\%, Code Phase Position (HDOP $>$ 3) 
		\\
		
		
		\cite{Stern2018} & 
		2016 & 
		5.71 hour (613 km)  & 
		GPS, GLO & 
		L1 & 
		Mass Market (+INS) & 
		SBAS & 
		Highway & 
		2.6 m, 95\%, Horizontal	& 
		100\% with inertial & 
		-
		\\
		
		\cite{Reid2019a} & 
		2018 & 
		355 hours (30,000 km) & 
		GPS, GLO & 
		L1, L2	& 
		Survey (+ INS) & 
		Net. RTK & 
		Mostly Highway & 
		1.05m, 95\%, Horizontal & 
		50\%~Integer Ambiguity Fixed	& 
		10 sec, 50\%, 40 sec, 80\%
		Fixed
		\\
		
        \cite{Humphreys2019} & 
        2019 & 
        2 hours & 
        GPS, Gal & 
        L1, L2	&
        Research SDR & 
        Net. RTK & 
        Urban & 
        0.14m, 95\%, Horizontal & 
        87\% Integer Ambiguity Fixed &
        2 sec, 99\%, Fixed
        \\

        Swift Navigation & 
        2019 & 
        12 hours (1,312 km) & 
        GPS, Gal & 
        L1, L2	& 
        Mid-Range & 
        Proprietary, Continent-Scale & 
        Mostly Highway & 
        0.35m, 95\%, Horizontal & 
        95\% CDGNSS &
        -\textbf{}
        \\

		\bottomrule
	\end{tabular}
	\\ 			
	\rule{0pt}{2ex}  
\end{table*}

The resulting GNSS performance increases of all the aforementioned development is demonstrated through select investigations in Table \ref{tab:gnss-perf-sum}. The important result is that, as of 2020, production GNSS systems with continent-scale corrections coverage can deliver 0.35m accuracy with availability at 95\%, and research systems reach 0.14m accuracy even in light urban scenarios. 

\subsection{Performance of a State-of-the-Art Production System}

In 2019, Swift Navigation performed an on-road performance assessment of a state-of-the-art production GNSS system using the methodology outlined by Reid et al~\cite{Reid2019a}. The positioning engine under test was Swift Navigation's Skylark™, running on a Piksi\textsuperscript{®} Multi GNSS receiver, equipped with a Harxon antenna. The GNSS corrections were provided by Swift Navigation's Skylark cloud corrections service, currently available across the contiguous United States. Ground truth was derived from a NovAtel SPAN GNSS-Inertial system~\cite{Kennedy2006}. Corrections were delivered over the 4G LTE cellular network. This setup was driven over 1,300~km from downtown Seattle, WA to downtown San Francisco, CA, along  the U.S. Interstate Freeway system. 

The results from this data collection campaign are shown in Table \ref{tab:gnss-perf-sum}. The 95th percentile accuracy performance is 0.35~m, with an availability of 95\%. These results represent state-of-the-art performance for a commercially available automotive grade GNSS positioning system. 

\subsection{Lane Determination with GNSS and Maps}

The automotive community is broadly interested in localization technologies that can determine which lane a vehicle is travelling in. Is the performance of modern GNSS good enough to reliably provide lane-determination capabilities? To answer this, we must account for both the error from the GNSS system and the map as follows: 

\begin{equation}
    \sigma_{GNSS}^2 + \sigma_{map}^2 = \sigma_{total}^2
\end{equation}
where $\sigma_{GNSS}$ is the standard deviation of the GNSS position, $\sigma_{map}$ is that for the HD map, and $\sigma_{total}$ is the total budget between them. 

Following the methodology presented in~\cite{Reid2019c}, it can be shown that the lateral position error budget for highway lane determination is 1.62~m for passenger vehicles in the U.S. For safe operation, it is recommended by~\cite{Reid2019c} that this position protect level be maintained to an integrity risk of 10\textsuperscript{-8} / h, or a reliability of 5.73$\sigma$ assuming a Gaussian distribution of errors. This gives us the following relationship: 

\begin{equation}
    5.73~\sigma_{total} < 1.62 \text{ m}
\end{equation}
solving for $\sigma_{total}$ gives: 
\begin{equation}
    \sigma_{total} < \frac{1.62 \text{ m}}{5.73}=0.28\text{ m}
\end{equation}

If we allow equal error budget for the GNSS and map georeferencing, $\sigma_{map} = \sigma_{GNSS} = \sigma_{alloc}$, then we obtain the following: 

\begin{equation}
    2 ~\sigma_{alloc}^2 = \sigma_{total}^2
\end{equation}
solving for this allocation for highway geometry gives: 
\begin{equation}
    \sigma_{alloc} = \frac{\sigma_{total}}{\sqrt{2}}=\frac{0.28\text{ m}}{\sqrt{2}}=0.20\text{ m}
\end{equation}

This allows us to calculate an approximate value for 95\% accuracy (1.96~$\sigma_{alloc}$) requirements for both the GNSS position and map georeferencing to be 1.96~$\sigma_{alloc}$~=~0.39~m. Given the performance of production-ready systems shown in Table \ref{tab:gnss-perf-sum}, we conclude that modern GNSS is accurate enough to provide lane-determination with high confidence, within reach of safety-of-life requirements. Simultaneously, modern HD maps and mapping techniques have also been shown to achieve the required accuracy~\cite{TheSanbornMapCompany2017, Dannehy2016, Slovick2019}.

\subsection{Addressing Availability Limitations}

Modern GNSS still suffers from outages which prevent stand-alone usage for fully autonomous applications where even short localization outages can result in mission failure. For this reason, GNSS is often aided with inertial and wheel, visual, and radar odometry inputs. These inputs suffer from long-term position drift. Fortunately, most GNSS outages during driving are only a few seconds long, as shown in Table \ref{tab:gnss-perf-sum}. Thus, a modern automotive INS-aided GNSS system might soon be able to achieve upwards of 99.9\% availability on highways. Urban environments remains challenging~\cite{Stern2018}.

\section{Emerging Autonomous Driving Architectures}
\label{sec:AV-arch}

Two architectures for autonomous vehicle are emerging. They differ in their sensor suites, driving capabilities, and intended Operational Design Domain (ODD). The first provides SAE Level 2 advanced driver-assistance on limited access roads in consumer vehicles, and are currently available to the public. The second provides SAE Level 4 driverless vehicle operation, such as those in robo-taxi platforms targeted at ride sharing, particularly within cities, and are currently in limited pilot operations. For investigations into different sensor choices for autonomous driving, we refer the reader to ~\cite{VanBrummelen2018, Rosique2019}. Here, we investigate both architectures with an eye on their localization approach and challenges.

\subsection{SAE Level 2 Vision-based Systems}

\begin{figure}[h]
    \centering
    \includegraphics[width=3.4in]{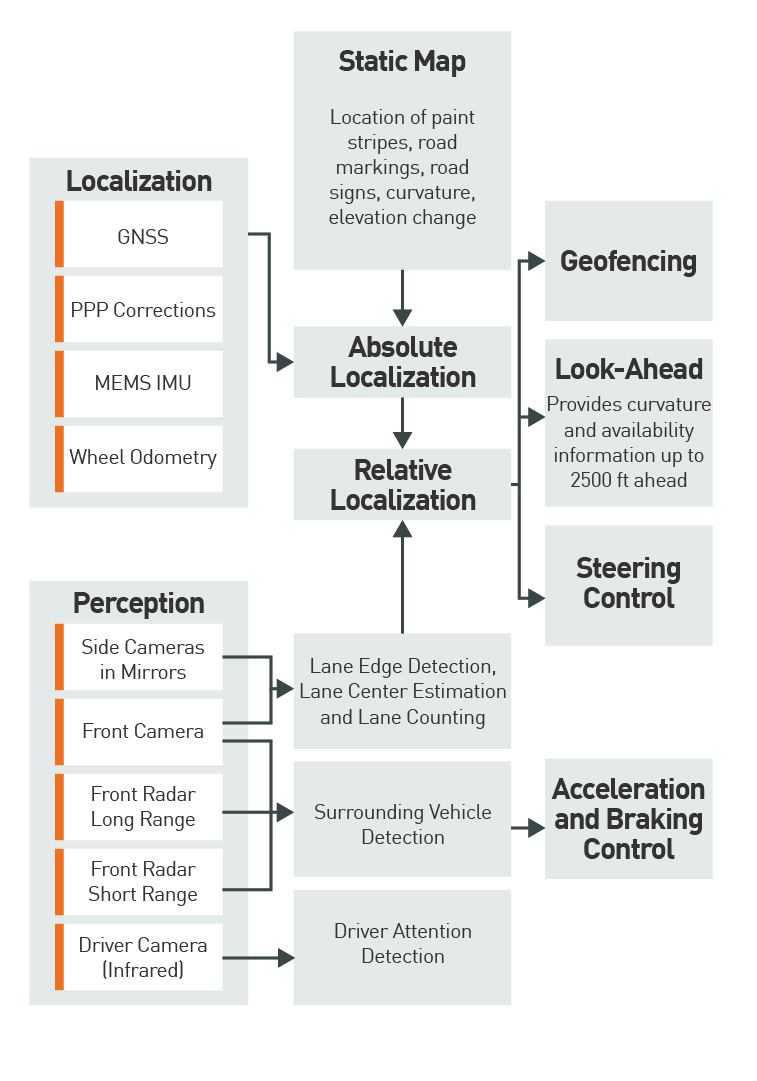}
  \caption{Common elements of traditional SAE Level 2 automated driving architectures for lane-following. A perception system provides lateral and longitudinal in-lane localization and detects surrounding vehicles. A dedicated localization and mapping system provides oversight over the perception system and enables planning beyond perception limits, such as slowing for upcoming curves. The desire to perform more complex maneuvers such as lane changes is evolving this architecture towards unifying precision GNSS for lane-level localization and camera-based in-lane localization to plan and execute paths.}
  \label{fig:L2-anatomy}
\end{figure}

An archetypal Level 2 architecture for autonomous lane-following under human supervision is shown in Figure \ref{fig:L2-anatomy}. Perception is used for in-lane control to keep the vehicle between lane lines without aid from maps and localization~\cite{BarHillel2014}. State-of-the-art Level 2 systems aim to move beyond lane-following and provide onramp-to-offramp freeway navigation, which requires selecting and changing into the correct lanes to traverse interchanges and merges and selecting or avoid exit lanes. This functionality was first demonstrated by Tesla's `navigate on autopilot' feature using computer vision and radar~\cite{TheTeslaTeam2018, TheTeslaTeam2019}. The Cadillac Super Cruise approach also utilizes computer vision and radar, but further employs precision GNSS and HD maps to (1) geofence the system to limited access divided highways and (2) provide extended situational awareness beyond perception range~\cite{Hay2018a,Davies2018}. 

The dependency of these Level 2 architectures on camera data leads to three major challenges. One, ambiguous road markings can lead such systems astray, steering vehicles into phantom lanes and potentially causing fatal accidents. Two, for lane-level maneuvers such as lane changes, navigating interchanges and merges, and choosing or avoiding exit lanes, the vehicle has to correctly infer its surrounding lanes and read, associate, and remember road signs to understand the lane's intended use. Lastly, these systems have no fallback in the face of environmental effects that can degrade, occlude or damage cameras. These challenges makes it difficult to reach the reliability required for safety-of-life deployment. Precision GNSS providing lane-level localization and lane determination, coupled with HD Maps, can address each of these challenges particularly well in highway environments.

\subsection{SAE Level 4 LiDAR-based systems}
\label{sec:sae-level-4}

\begin{figure}[h]
    \centering
    \includegraphics[width=3.3in]{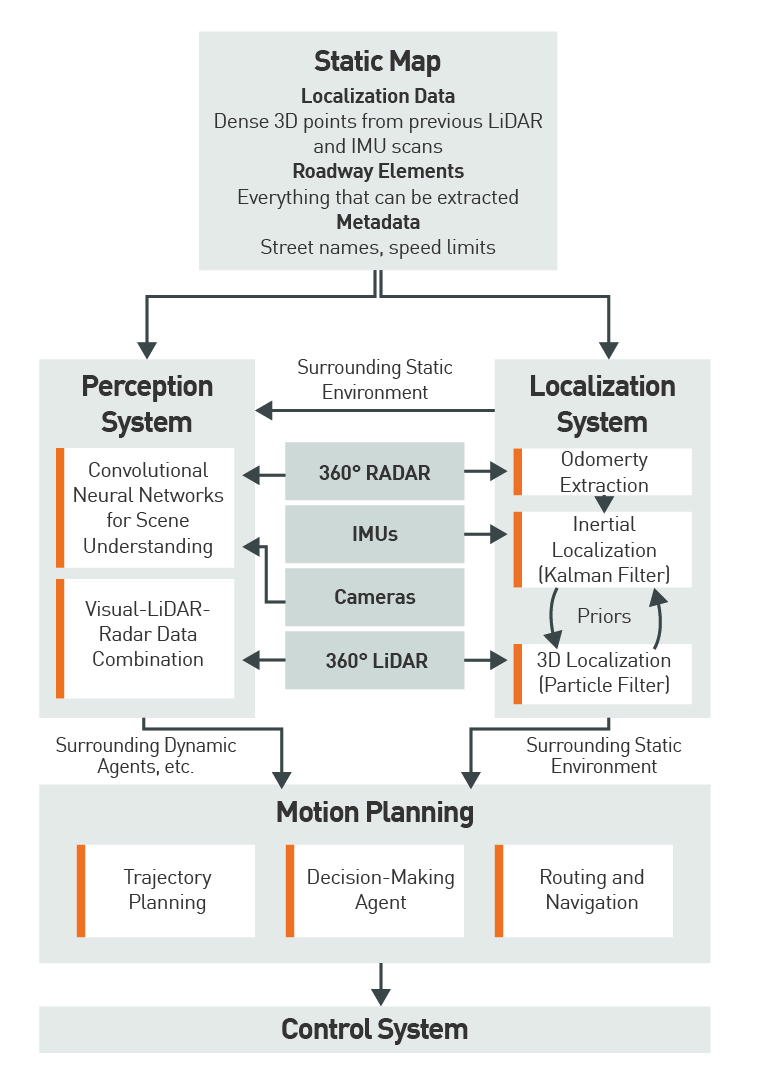}
  \caption{Common elements of SAE Level 4 automated driving architectures. Sensor data flows from LiDAR, cameras, RADAR, IMUs, GNSS and others to both the localization and perception system. The localization system tracks the vehicle's pose by fusing relative motion from inertial, wheel, and possibly radar data with map-relative localization. The localization and mapping system provides enough fidelity to solve the driving problem in static environments, freeing perception system to focus on detecting dynamics in the environment, such as moving actors, traffic light states, and roadwork. A representation of the environment containing both the surrounding static map from localization and the dynamic elements from perception is passed to motion planning, which hierarchically solves for the path the vehicle will follow. }
  \label{fig:L4-anatomy}
\end{figure}

Level 4 systems under development intend to fully automate the dynamic driving tasks within its ODD, with no vehicle operator required. The archetypal architecture for Level 4 driving is given in Figure \ref{fig:L4-anatomy}. One of the insights shared by most Level 4 systems is to simplify the driving problem through high accuracy maps and localization. Indeed, most Level 4 systems cannot function if the localization or mapping subsystem is unavailable, and localization failures often trigger an emergency stop. 

Level 4 architectures have historically relied on LiDAR for localization, achieving \textless~0.10~m, 95\% lateral and longitudinal positioning accuracy~\cite{Liu2019}. LiDAR localization approaches can leverage 3D structure~\cite{Segal2009a} and surface reflectivity\cite{Levinson, Wolcott2015, Wolcott2016b}. GNSS is not the primary localization sensor due in part to historical availability challenges~\cite{Urmson2004}. Unfortunately, LiDAR also suffers from outages, faults, and erroneous position outputs, especially during adverse weather or in open featureless environments such as highways~\cite{Sundararajan2016, Michaud2015, Kutila2016, Phillips2017, Heinzler2019}. Inertial and odometry-based navigation is commonly used in conjunction with LiDAR to address some of these concerns, but suffer from position drift over time. An additional absolute localization sensor would aid in providing redundancy to overcome outages and detect LiDAR errors and faults.

Precision GNSS is complementary to LiDAR. GNSS' microwave signals are unaffected by rain, snow, and fog. GNSS also performs best in open sparse environments like highways. For these reason, precision GNSS can aid LiDAR-based localization to reach safety-of-life levels of reliability and coverage. One example of a Level 4 system that leverages precision GNSS is Baidu's Apollo framework~\cite{Wan2018}.

\section{Discussion}

We have hinted at the potential benefits of GNSS given the current localization challenges for autonomous driving architectures. We discuss each benefit here.

\subsection{Unlocking Lane-Level Maneuvers}

We have shown that GNSS can provide lane determination with safety-of-life level integrity, especially valuable to Level 2 vision-based systems given their challenges outlined previously. Lane determination on an HD map enables planning lane-level maneuvers with confidence, providing a key building block for safe onramp-to-offramp navigation.

\subsection{Providing Oversight over Vision Systems}

Vision systems can be fooled into detecting phantom lanes leading to extremely hazardous behavior. GNSS lane determination provides an independent signal to verify the validity of vision outputs.

\subsection{Providing Safety Through Independence}

The challenge facing automakers is achieving reliability of better than 10\textsuperscript{-8} dangerous failures per hour, exceeding ASIL D~\cite{Kafka2012}. This allowable failure rate has been estimated as less than one failure in a billion miles~\cite{Reid2019c}. One powerful method to reach this level of reliability is combining independent systems to redundantly localize the vehicle. GNSS can provide this independent signal.

\subsection{Fallback During Outages}

Both LiDAR and vision systems experience outages. Level 4 systems rely on expensive tactical-grade Inertial Measurement Units (IMU) to safely pull over in these circumstances, and Level 2 systems depend on the driver to intervene. Precision GNSS might be a viable fallback during outages, enabling operation in a degraded mode for Level 4 systems and aiding in advanced warning of a required driver intervention to transform Level 2 systems into Level 3 systems.

\subsection{Unlocking Interoperability to Overcome Occlusion and Enable Collaboration}

An important future opportunity is interopreability between autonomous systems, which cannot be understated. For one, information sharing between vehicles and static infrastructure is a powerful approach to overcoming perceptual sensor occlusion --- a major challenge for autonomous vehicles. Furthermore, interoperability enables strategic and tactical collaboration, augmenting basic driving to unlock coordination between vehicles and commodifying enabling technologies such as HD Maps. Interoperability requires sharing common spatial reference frames and timing. The obvious choice for a global standard is that defined by GNSS, namely, the ITRF datum and GPS global time. GNSS offers the only source of globally consistent precise position and time to act as a standard reference for all autonomous systems. 

\bibliographystyle{IEEEtran}
\bibliography{main}

\end{document}